\documentclass[letterpaper, 10 pt, conference]{ieeeconf}  

\IEEEoverridecommandlockouts                              

\usepackage{graphicx} 
\usepackage{amsmath} 
\usepackage{amssymb}  
\usepackage{bm}  
\usepackage{xcolor}
\usepackage{lipsum}

\usepackage{flushend}

\usepackage{amsthm}

\newtheorem{assumption}{Assumption}

\title{\LARGE \bf
Distributed Control Barrier Functions for Safe Multi-Vehicle Navigation in Heterogeneous USV Fleets
}

\author{Tyler M. Paine$^{1}$, Brendan Long$^{2}$, Jeremy Wenger$^{1,3,4}$, Michael DeFilippo$^{1}$, James Usevitch$^{5}$, Michael R. Benjamin$^{1}$
\thanks{$^{1}$Department of Mechanical Engineering, Massachusetts Institute of Technology, 
       Cambridge, MA 02139, USA
       {\tt\small tpaine@mit.edu, jwenger@mit.edu, mikedef@mit.edu, mikerb@mit.edu}}%
\thanks{$^{2}$MIT Lincoln Laboratory, Lexington, MA 02421, USA
       {\tt\small brendan.long@ll.mit.edu}}%
\thanks{$^{3}$Woods Hole Oceanographic Institution
       Woods Hole, MA 02543, USA}%
\thanks{$^{4}$Charles Stark Draper Laboratory
       Cambridge, MA 02139, USA}%
\thanks{$^{5}$Department of Electrical and Computer Engineering,                 Brigham Young University,
       Provo, UT 84602, USA}%
\thanks{\scriptsize DISTRIBUTION STATEMENT A. Approved for public release. Distribution is unlimited. This material is based upon work supported by the Under Secretary of War for Research and Engineering under Air Force Contract No. FA8702-15-D-0001 or FA8702-25-D-B002. Any opinions, findings, conclusions or recommendations expressed in this material are those of the author(s) and do not necessarily reflect the views of the Under Secretary of War for Research and Engineering. © 2026 Massachusetts Institute of Technology. The portions of this work supported by the U.S. Government are delivered to the U.S. Government with Unlimited Rights, as defined in DFARS Part 252.227-7013 or 7014 (Feb 2014). Notwithstanding any copyright notice, U.S. Government rights in those portions of this work are defined by DFARS 252.227-7013 or DFARS 252.227-7014 as detailed above. Use of those portions of this work other than as specifically authorized by the U.S. Government may violate any copyrights that exist in this work}
}

\begin{document}

\maketitle
\thispagestyle{empty}
\pagestyle{empty}

\begin{abstract}
Collision avoidance in heterogeneous fleets of uncrewed vessels is challenging  because the decision-making processes and controllers often differ between platforms, and it is further complicated by the limitations on sharing trajectories and control values in real-time. 
This paper presents a pragmatic approach that addresses these issues by adding a control filter on each autonomous vehicle that assumes worst-case behavior from other contacts, including crewed vessels.  This distributed safety control filter is developed using control barrier function (CBF) theory and the application is clearly described to ensure explainability of these safety-critical methods.    
This work compares the worst-case CBF approach with a Collision Regulations (COLREGS) behavior-based approach in simulated encounters.   
Real-world experiments with three different uncrewed vessels and a human operated vessel were performed to confirm the approach is effective across a range of platforms and is robust to uncooperative behavior from human operators.
Results show that combining both CBF methods and COLREGS behaviors achieves the best safety and efficiency. 

\end{abstract}

\section{INTRODUCTION}

\begin{figure}[ht]
    \centering
    \includegraphics[width=1.0\linewidth]{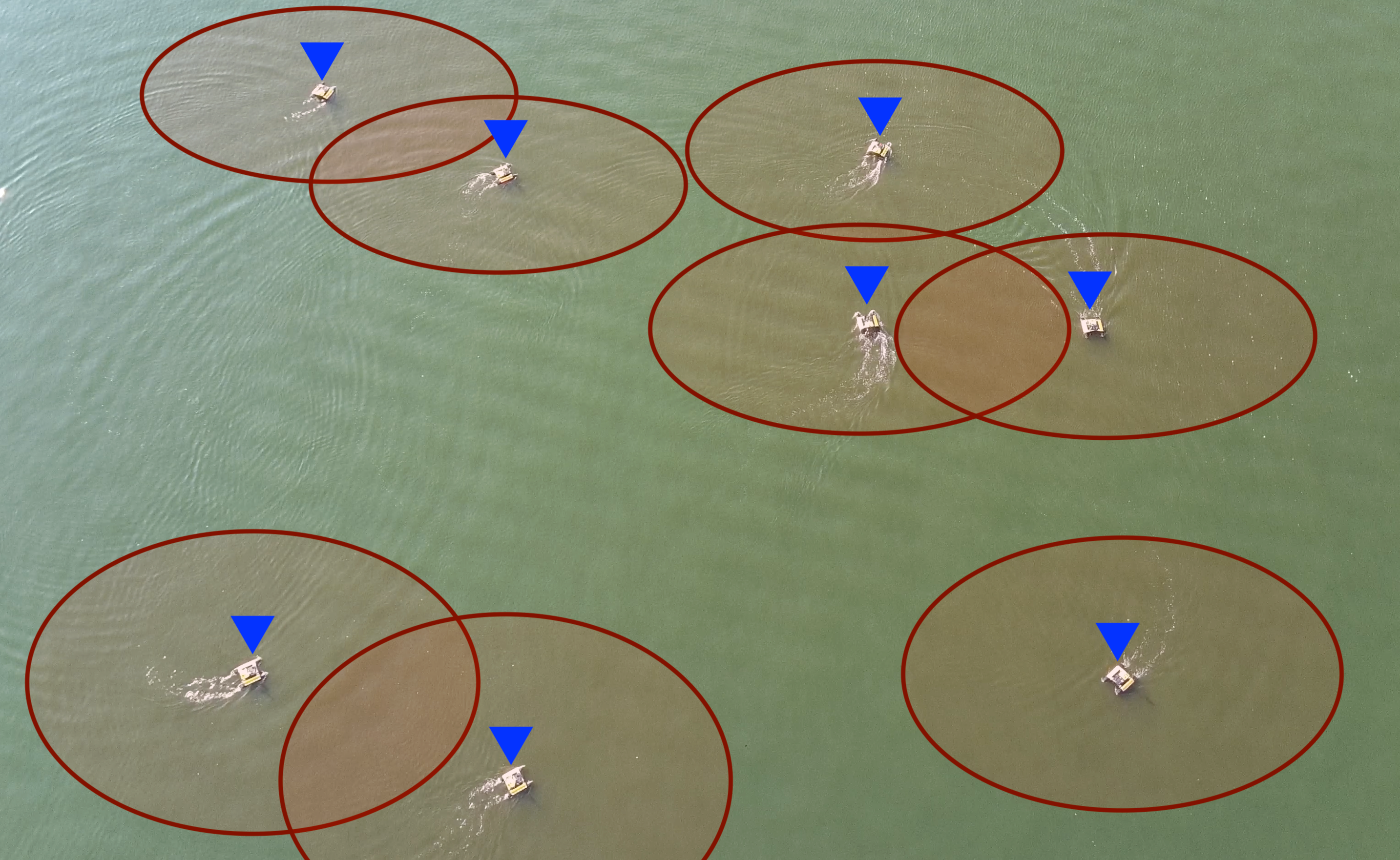}
    \caption{Fleet of USVs using the distributed control barrier function (CBF) method described in this paper while performing a search mission on the Charles River.  A graphical depiction of the barriers are shown by the red circles. }
    \label{fig:USV_fleet_exp}
    \vspace{-3mm}
\end{figure}

The number of uncrewed surface vehicles (USVs) is increasing, with the goal of using them in large cooperative teams so they can complete tasks more efficiently and robustly \cite{CRS_DIU_Report}.  An important part of realizing the benefits of collaboration is ensuring that autonomous USVs do not collide with other vessels,  even when other vessels are not perfectly following a collision protocol themselves.  This objective is not easy since USVs often need to operate in close proximity to other USVs and crewed vessels when performing collaborative tasks such as formation following \cite{turrisi2024IROS}, collaborative searching \cite{Gershfeld2023RaL}, or high value unit (HVU) defense \cite{OPNAV2020HVU}.

Although autonomous vehicles can be programmed with the \textit{intention} of avoiding collisions, there is still a gap between desired behavior and what can be executed in complex situations and real environments.  This difference can lead to unexpected `in extremis' situations \cite{USCG_NavRules_Handbook_2024} where a collision can be only avoided with the correct emergency maneuvers.  
These situations come up periodically in multi-vehicle missions, even when individuals are using policies or behaviors that follow a collision avoidance protocol.  


There are two main causes for this gap.  First, issues arise when multiple mission objectives are active at the same time, making it difficult to find a safe control input simultaneously satisfying all objectives.  The root of this problem is that COLREGS \cite{USCG_NavRules_Handbook_2024}, the rules of the road for marine vehicles that are the basis for many autonomous behaviors or policies, only specifies how to maneuver when encountering a \textit{single} contact, not multiple contacts.  When more than one contact is present, multiple single-contact behaviors are used simultaneously.  Moreover, these collision avoidance behaviors are usually combined with other behaviors related to mission objectives within the context of a multi-objective optimization problem. 
Although each behavior is designed to produce rational utility functions, these functions are repeatedly combined with others during a mission.  In complex situations it can be challenging to determine the optimal weighting of this mixture at all times, and as a result sometimes the net outcome does not adequately prioritize safety.  
Secondly, many COLREGS-based navigation algorithms operate by updating vehicles' planned trajectories in response to the presence of external vehicles.  However, there can exist a delay between updating a trajectory and low-level control algorithms successfully tracking the new trajectory. 
Even if the new trajectory may be safe, the closed-loop response may be too slow to safely avoid imminent collisions.

These issues can be addressed by the use of control barrier functions (CBFs) \cite{ames2019control, Garg2024AnnualRev}.  CBFs are a well-established class of control filtering methods that minimally modify nominal control inputs to ensure satisfaction of safety constraints. Prior literature has demonstrated that CBFs exhibit excellent online performance for maintaining safety in scenarios involving multiple agents and/or unanticipated obstacles due to their ability to simultaneously compose and enforce multiple set invariance constraints. In scenarios where the intent or control actions of other vehicles are not known to the ego vehicle, adversarial control barrier functions \cite{usevitch2022adversarial} can be used to enforce safety conditions under the assumption of worst-case control actions from external vehicles.

The focus of this paper is to report the application of recently developed theoretical results for control filtering to the specific problem of safely operating groups of USVs in completely autonomous fleets and alongside crewed vessels.  The specific contributions include:
\begin{itemize}
    \item Addressing the problem of `in extremis' collision avoidance for autonomous marine vehicles using distributed CBF methods that do not require information about intent or control to be shared. 
    \item Evaluation of the method in simulation and experimental trials with a fleet of heterogeneous USVs and a crewed vessel. 
\end{itemize}

\section{RELATED WORK}

COLREGS collision avoidance protocol essentially prescribes a bias applied
to the anticipated closest point of approach (CPA) for a candidate
maneuver, e.g, each vessel should pass the other on the port side in a
head-on situation. At some point of imminent danger or proximity, the
in-extremis mode of the COLREGS is engaged and the obligation is then to
perform whatever shiphandling necessary to avoid collision. A quote often
attributed to Admiral Ernest King of the U.S. Navy is "The mark of a great
shiphandler is never getting into situations that require great
shiphandling.". Several methods have been put in to practice for COLREGS
maneuvering, with the goal of never reaching an in-extremis situation. The
Velocity Obstacle approach \cite{kuwata2010} defines a cone of collission-free
maneuvers from which maneuver selection is a constraint satisfaction
problem. Rapidly-exploring Random tree (RRT) based COLREGS path planning,
e.g., \cite{chiang2018}, has been used for quickly finding collision free paths
in marine traffic problems. In this work, the IvP Helm is used by forming
a multi-objective optimization (IvP) problem with functions generated by
distinct behaviors, each correlated to a given contact \cite{benjamin2006ColregsBhv}. This method allows
collision avoidance to be considered more generally than a binary
constraint satisfaction problem. IvP functions can also capture maneuvers
that are less than ideal, which is critical in identifying compromise
maneuvers in multi-vehicle COLREGS situations.

Control barrier functions (CBFs) have theoretical roots stemming from Nagumo's necessary and sufficient conditions for set invariance \cite{menner2024translation, nagumo1942lage}, optimization interior point methods \cite{gondzio2012interior}, barrier certificates in hybrid systems \cite{prajna2004safety}, and control Lyapunov functions \cite{sontag1989universal, lin1991universal}. A conservative form of CBFs was first introduced in \cite{wieland2007constructive}, but their modern form was popularized in \cite{ames2016control}. CBFs were originally considered only for single-agent systems, but later work presented extensions to multi-agent systems \cite{wang2017safety, wang2016safety, wang2016multi, zhang2025gcbf+}. However, a common assumption among these early works on multi-agent CBFs is that all agents follow the nominally specified control law and cooperate to ensure collision-free operations. This assumption was removed in \cite{usevitch2021adversarial, usevitch2022adversarial}, which introduced the notion of adversarially resilient control barrier functions that consider worst-case behavior of neighboring agents attempting to violate safety conditions.
Readers interested in a more thorough review of the history and applications of CBFs are referred to \cite{ames2019control, Garg2024AnnualRev}.

Prior work most similar to this paper includes \cite{lee2025turningcirclebasedcontrolbarrier} and \cite{Thyri2020CBF_COLREGS}. The work \cite{lee2025turningcirclebasedcontrolbarrier} considers novel CBFs for underactuated, non-holonomic autonomous surface vehicles. However, it does not consider either cooperative or worst-case multi-agent interactions. The work \cite{Thyri2020CBF_COLREGS} uses CBFs to ensure that the target ship (TS) domain defined by the COLREGS is not violated by the ego vehicle. However, this work does not consider attempting to satisfy COLREGS specifications under worst-case misbehavior from neighboring agents.

\section{PRELIMINARIES}
\subsection{Notation and Definitions}
Vectors are in bold, i.e.  $\bm{x} \in {\rm I\!R}^n$ and matrices are capitalized in bold, i.e. $\bm{A} \in {\rm I\!R}^{n \times n}$. 
The partial Lie derivative of $h(\bm{x})$ with respect to $\bm{x}$ along a function $g$ is $L_g h^{\bm{x}}(\bm{x}) = \frac{\partial h (\bm{x})}{\partial \bm{x}} \cdot g(\bm{x})$. 

There are $N_a$ vehicles contained within the set $\mathcal{V}$ and indexed $\{ 1, 2, \dots N_a\}$. 
The index of the ego vehicle is $i$, while the index of other vehicles are denoted as $j$.

\subsection{Vehicle State and Dynamics Model}
\begin{figure}[ht]
    \centering
    \includegraphics[width=0.6\linewidth]{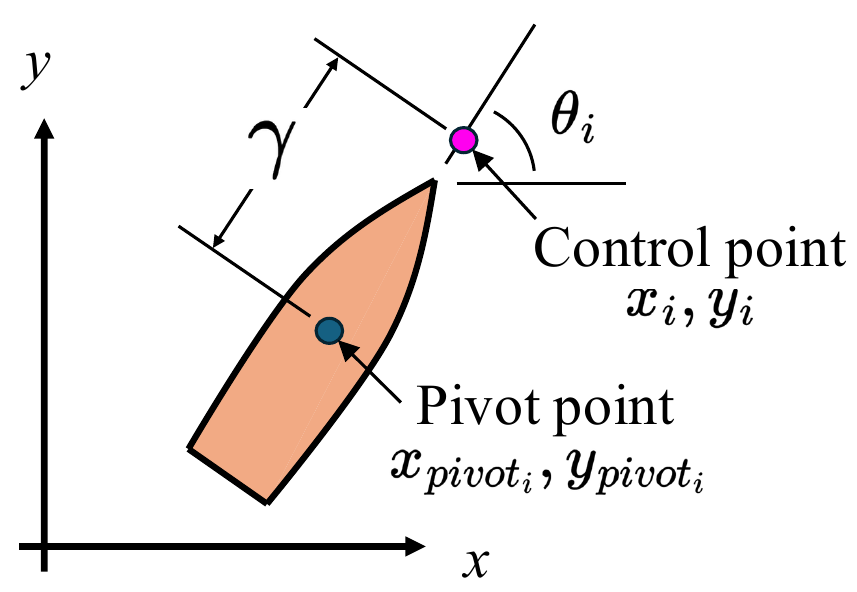}
    \caption{State definition for the $i^{th}$ or ego vehicle}
    \label{fig:state_def}
    \vspace{-3mm}
\end{figure}

The control point of the surface vehicle in the global frame is represented by 
\begin{align} \label{eq:control_state}
    \bm{x}_i = \begin{bmatrix}
        x_i \\ y_i \\ \theta_i
    \end{bmatrix} \in {\rm I\!R}^2 \times \mathcal{S}
\end{align}
where the index $i$ denotes the $i^{th}$ vehicle.  As shown in Figure \ref{fig:state_def}, the control point is a hypothetical point located a distance $\gamma$ in front of the pivot point of the vehicle.  The classic unicycle dynamics are recovered when $\gamma = 0$. 

The dynamics of each agent is modeled as a first-order nonlinear system
\begin{equation}
    \dot{\bm{x}}_i = g(\bm{x}_i)\bm{u}_i,  \label{eq:SI_dynamics}
\end{equation}
where $g(\bm{x}_i)$ is locally Lipschitz, and the control input
\begin{align}
    \bm{u}_i = \begin{bmatrix}
        u_{thr} \\ u_{rud}
    \end{bmatrix} \in {\rm I\!R}^2
\end{align}
where $u_{thr}$ is the commanded thrust and $u_{rud}$ is the commanded rudder.   It is assumed that the control input constraints are represented by a nonempty, convex, compact polytope $\mathcal{U}$ that are symmetric about the origin \cite{usevitch2021adversarial}.  
For the types of surface vehicles considered in this paper, the requirement is represented by 
\begin{align}
    \mathcal{U} = \{ \bm{u} \subset {\rm I\!R}^2 : -u_{thr_{min}} \leq& \ u_{thr}  \leq u_{thr_{max}}, \\
     -u_{rud_{min}} \leq& \  u_{rud}  \leq u_{rud_{max}} \}.
\end{align}

Under a nominal feedback controller $\bm{u}_i = k(\bm{x}_i) \in {\rm I\!R}^2$ the closed loop dynamics of an agent's state are modeled as
\begin{equation}
    \dot{\bm{x}_i} = g(\bm{x}_i)k(\bm{x}_i).  \label{eq:cbf_dynamics}
\end{equation}

The single-order model \eqref{eq:cbf_dynamics} is used in this approach because of the assumptions about communication and cooperation.  More specifically, it is assumed that the second-order dynamics and the control input of other vehicles are not known, rendering higher-order models useless in this context.  
Finally, three key assumptions regarding the information required to use this method are provided. 

\begin{assumption}\label{assump:state_known}
The neighbor's state $\bm{x}_j$ is known to the $i^{th}$, or ego vehicle. 
\end{assumption}

\begin{assumption}\label{assump:ctrl_set_known}
The neighbor's control input set $\mathcal{U}_j$, or a conservative over-approximation of this set is known to the $i^{th}$, or ego vehicle.    Given the first-order model of dynamics \eqref{eq:SI_dynamics}, this assumption is equivalent to the ego vehicle knowing the max speed and yaw-rate of other vehicles. 
\end{assumption}

\begin{assumption}\label{assump:ith_control_set}
Using its control inputs $\bm{u}_i \in \mathcal{U}_i$ the $i^{th}$ vehicle can drive its state such that $\dot{x}_i \geq \dot{x}_j$ and $\dot{y}_i \geq \dot{y}_j$.  In other words, the ego vehicle can move its control point as fast or faster than the neighbor vehicle can move theirs.  
\end{assumption}

\subsection{Control Barrier Functions for Control Filtering}
USVs are safe if they remain inside a safe set, or within safe areas on the surface, a subset of ${\rm I\!R}^2$. 
This safe set is formally defined by $\mathcal{C}$, a superlevel set of a continuously differentiable function $h: D \subset {\rm I\!R}^n \to {\rm I\!R}$ such that 
\begin{align}
    \mathcal{C} =& \{ \bm{x}_i \in D \subset {\rm I\!R}^n : h(\bm{x}_i) \geq 0 \}, \label{eq:p1}\\
    \partial\mathcal{C} =& \{ \bm{x}_i \in D \subset {\rm I\!R}^n : h(\bm{x}_i) = 0 \}, \label{eq:p2}\\
    \text{Int}(\mathcal{C}) =& \{ \bm{x}_i \in D \subset {\rm I\!R}^n : h(\bm{x}_i) > 0 \}, \label{eq:p3}
\end{align}
The set $\mathcal{C}$ is forward invariant if for every $\bm{x}_i(0) \in \mathcal{C}$, the evolution of the system \eqref{eq:cbf_dynamics} $\bm{x}_i(t) \in C \ \forall \ t$.  See \cite{ames2016control} for more details. 

\begin{figure}[ht]
    \centering
    \includegraphics[width=1.0\linewidth]{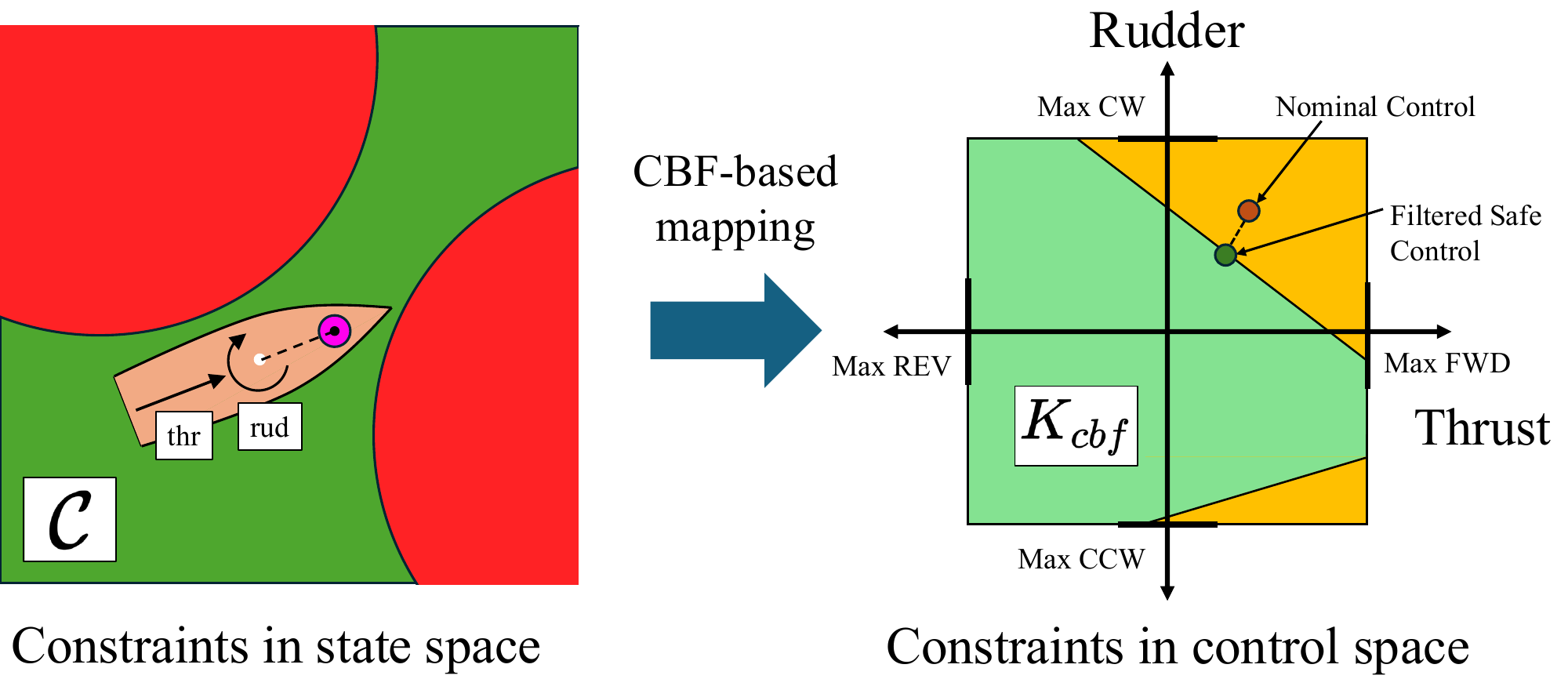}
    \caption{Example of mapping of constraints in state space to control input space and the process of control filtering.  Left: a vehicle must remain within the safe set $\mathcal{C}$ marked as the green areas.  This state constraint is translated into constraints in the control space, rendering some combinations of the rudder and thrust unsafe.  The nominal control is unsafe and is filtered to the closest safe control input. }
    \label{fig:cbf_relationship}
    \vspace{-3mm}
\end{figure}

From \cite{ames2016control}, the set consisting of all control values where the set $\mathcal{C}$ is forward invariant for the system \eqref{eq:cbf_dynamics} is defined as
\begin{equation}
    K_{cbf}(\bm{x}_i) = \{ u_i \in \mathcal{U} : L_{g} h^{\bm{x}_i}  (\bm{x}_i) \bm{u}_i + \alpha ( h(\bm{x}_i)) \geq 0 \}
\end{equation}
where $\alpha$ is an extended class $\mathcal{K}_{\infty}$ function \cite{ames2016control}.  An illustration of the set $K_{cbf}$ and its relationship to the safe set $\mathcal{C}$ is shown in Figure \ref{fig:cbf_relationship}.

\section{MAIN RESULT: THEORY}\label{sec:theory}
In a multi-vehicle mission each USV must independently restrict their control inputs to the set $K_{cbf}(\bm{x})$, rendering the safe set $\mathcal{C}$ forward invariant for all vehicles.   
This problem must be solved without knowledge of the control inputs of other agents, a practical constraint that ensures the solution method can be extended to large groups where some agents are not cooperating but safety is still needed. Since knowledge of other agents control actions is not accessible, each agent assumes every other agent uses the worst case control inputs.

The formulations of $\mathcal{C}$ and $K_{cbf}(\bm{x})$ that address the distributed multi-vehicle collision safety problem are developed in the remainder of this section.   Then the control filtering is designed as a quadratic program (QP) using the bounds of $K_{cbf}(\bm{x})$ as constraints on the solution.

\subsection{Control constraints for avoiding adversarial agents}\label{sec:ctrl_constraints}
Building upon the theoretical development introduced in \cite{usevitch2021adversarial}, we define the pairwise function between the $i^{th}$ and $j^{th}$ agents as
\begin{equation}
    h_{pair}(\bm{x}_i, \bm{x}_j),  \label{eq:h_pair}
\end{equation}
that has the properties listed in \eqref{eq:p1}-\eqref{eq:p3}.  The function \eqref{eq:h_pair} includes the state of another agent, $\bm{x}_j$, in addition to the state of the ego agent.  An example of such a function is provided in Section \ref{sec:implementation}. The central argument in \cite{ames2016control} is based on an application of Nagumo's theorem \cite{nagumo1942lage} that shows the set $\mathcal{C}$ is forward invariant if  $\dot{h}(\bm{x}_i, \bm{x}_j) \geq - \alpha( h(\bm{x}_i, \bm{x}_j))$.  Thus the control input from both agents play a role in maintaining safety. 

Expanding $\dot{h}_{pair}(\bm{x}_i, \bm{x}_j)$ yields
\begin{equation}
    L_{g}h_{pair_{ij}}^{\bm{x}_i}(\bm{x}_i)\bm{u}_i + L_{g}h_{pair}^{\bm{x}_j}(\bm{x}_j) \bm{u}_j \geq -\alpha(h_{pair}(\bm{x}_i, \bm{x_j})) \label{eq:cbf_const1}
\end{equation}

The key issue is that $\bm{u}_j$, the control input from other agent is unknown.  Instead, the ego agent must assume the other agent executes the worst-case control input.  The effect of this worst-case control input on $h_{pair}(\bm{x}_i, \bm{x}_j)$ is bounded by
\begin{equation}\label{eq:zeta_min_j}
    \zeta^{min}_j = \min_{\bm{u}_j \in \mathcal{U}_j} \ \  L_{g}h_{pair_{ij}}^{\bm{x}_j}(\bm{x}_j) \bm{u}_j
\end{equation}
As explained in \cite{usevitch2021adversarial}, this term is the lower bound on the other agent's worst case control effort to minimize the LHS of \eqref{eq:cbf_const1}.  The term $\zeta^{min}_j$ is computed locally by the $i^{th}$ agent by solving a parametric linear program provided Assumption \ref{assump:state_known} and Assumption \ref{assump:ctrl_set_known} are true. 

The effect of the worst case control effort can be substituted into \eqref{eq:cbf_const1} to yield
\begin{equation}
    L_{g}h_{pair_{ij}}^{\bm{x}_i}(\bm{x}_i)\bm{u}_i  + \zeta^{min}_j \geq -\alpha(h_{pair_{ij}}(\bm{x}_i, \bm{x_j})), \label{eq:cbf_const_with_zeta}
\end{equation}
a linear constraint on $\bm{u}_i$.   This \textit{CBF derivative condition} \cite{Garg2024AnnualRev} defines the set of safe control actions, 
\begin{align}
    K_{cbf_{pair}}&(\bm{x}_i, \bm{x}_j) = \{ u_i \in \mathcal{U} :  \\ &L_{g}h_{pair_{ij}}^{\bm{x}_i}(\bm{x}_i)\bm{u}_i + \zeta^{min}_j + \alpha(h_{pair}(\bm{x}_i, \bm{x_j})) \geq 0 \} \nonumber
\end{align}

Given Assumption \ref{assump:ith_control_set} is true, the ego agent can maintain the inequality in \eqref{eq:cbf_const_with_zeta} \cite{usevitch2021adversarial}.

\subsection{Distributed computation of safe set}
This section extends the development in Section \ref{sec:ctrl_constraints} to the scenario when the ego vehicle has more than one neighbor.  The approach is to combine the constraints that arise from multiple pairwise functions $h_{pair}(\bm{x}_i, \bm{x}_j)$, since finding a single function $h(\bm{x}_1, \bm{x}_2,  \ldots \bm{x}_{N_{a}} )$ that satisfies \eqref{eq:p1}-\eqref{eq:p3} is not trivial.  

\begin{figure}[ht]
    \centering
    \includegraphics[width=1.0\linewidth]{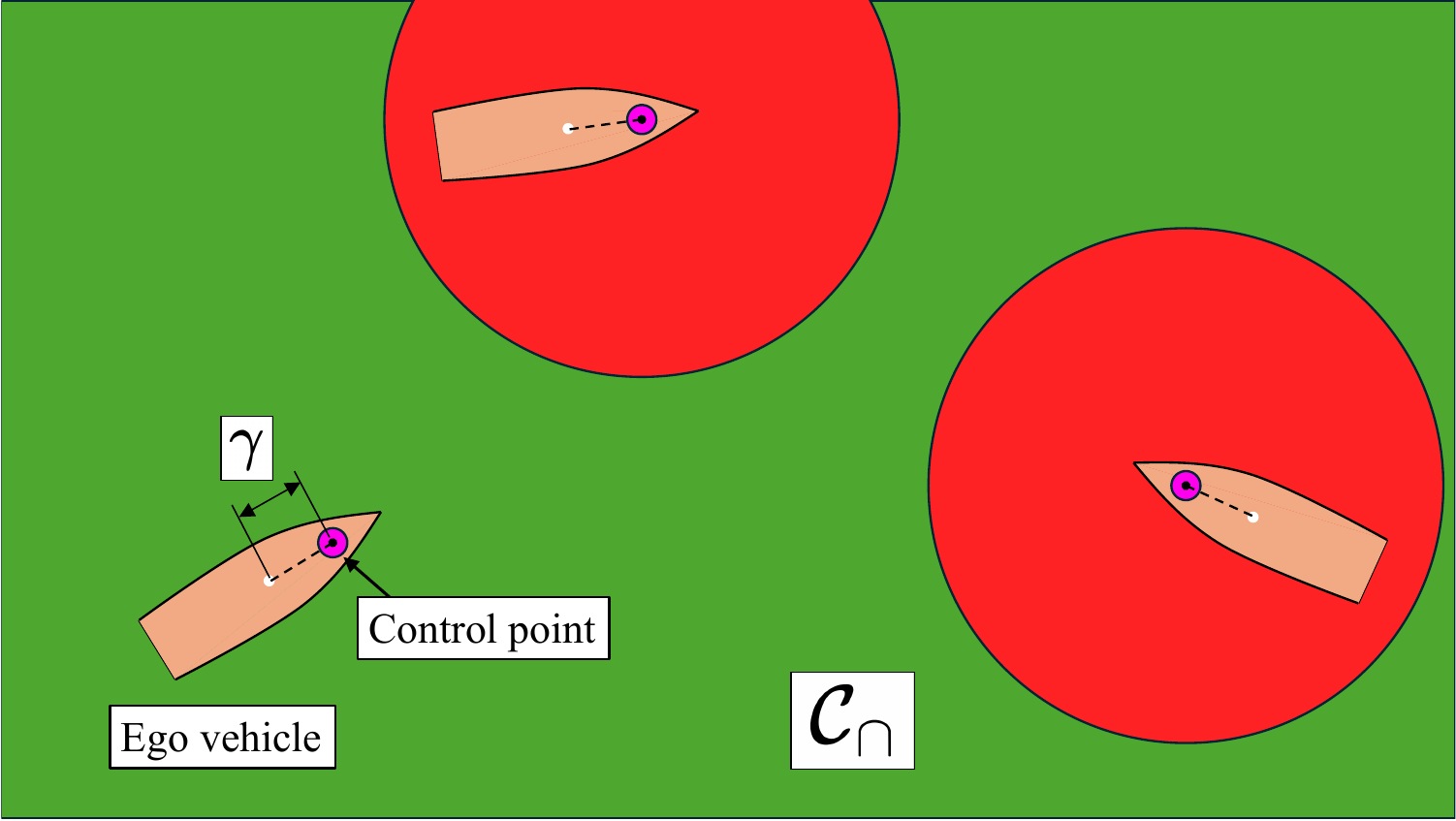}
    \caption{The combined safe set $\mathcal{C}_\cap$ (green) is the intersection of all the pairwise safe sets $\mathcal{C}$.  }
    \label{fig:combined_safe_set}
    \vspace{-5mm}
\end{figure}

The combined safe set of states $\mathcal{C}_\cap$ is the intersection of the safe sets that correspond to the pairwise barrier functions \eqref{eq:h_pair}, e.g.
\begin{equation}
    \mathcal{C}_{\cap_i} = \bigcap_{\substack{j\in\mathcal{V} \\ j \neq i}} \ \mathcal{C}_{h_{pair_{ij}}}, \label{eq:intersect_state_set}
\end{equation}
where the individual sets are denoted by $\mathcal{C}_{h_{pair_{ij}}}$ for clarity.  The set $\mathcal{C}_\cap$ is illustrated in Figure \ref{fig:combined_safe_set}. 

As a consequence of the definition of $\mathcal{C}_{\cap_i}$ in \cite{Molnar2023Composing}, the corresponding safe set of control inputs $K_{cbf_i}$ is an intersection of the corresponding individual sets $K_{cbf_{pair}}(\bm{x}_i, \bm{x}_j)$, e.g.
\begin{equation}
    K_{cbf_i} = \bigcap_{\substack{j\in\mathcal{V} \\ j \neq i}} K_{cbf_{pair}}(\bm{x}_i, \bm{x}_j), \label{eq:K_intersection}
\end{equation}
This set can be expressed as a combination of linear constraints on $\bm{u}_i$.  For example if $i=1$ this is
\begin{align}
    & K_{cbf_{1}}(\bm{x}_1, \bm{x}_2,  \ldots \bm{x}_{N_{a}} ) \\&= \{ u_1 \in \mathcal{U} :  \nonumber \\
    & \ \ \ \ \ \ L_{g}h_{pair_{ij}}^{\bm{x}_1}(\bm{x}_1)\bm{u}_1 + \zeta^{min}_2 + \alpha(h_{pair}(\bm{x}_1, \bm{x_2})) \geq 0 \nonumber \\
    & \ \ \ \ \ \ L_{g}h_{pair_{ij}}^{\bm{x}_1}(\bm{x}_1)\bm{u}_1 + \zeta^{min}_3 + \alpha(h_{pair}(\bm{x}_1, \bm{x_3})) \geq 0 \nonumber \\
   & \quad \quad \quad \quad \quad \quad \quad \quad \ \vdots  \nonumber \\
    & \ \ \ \ \ \ L_{g}h_{pair_{ij}}^{\bm{x}_1}(\bm{x}_1)\bm{u}_1 + \zeta^{min}_{N_a} + \alpha(h_{pair}(\bm{x}_1, \bm{x_{N_a}})) \geq 0\} \nonumber
\end{align}

\subsection{Formulation of QP for control filtering}
This section describes the process of using the result \eqref{eq:K_intersection} to design the control filter as a QP. 

The filtering process starts after another algorithm has computed a nominal control input, $\bm{u}_{nom_i}$, usually a function of mission objectives.  Then the goal of the filter is to compute a value for $\bm{u}_i$ that is within the set $K_{cbf_{i}}$ and as close as possible to $\bm{u}_{nom_i}$.  More formally, in the case of $j=2,3, \ldots N_a$ the QP is formulated as
\begin{align}
     \min_{u_i \in \mathcal{U}_i}& \ \ (\bm{u}_i - \bm{u}_{nom_i})^T \bm{Q} (\bm{u}_i - \bm{u}_{nom_i})\\
    \text{s.t.} &  
                   \begin{bmatrix}
                    -L_{g}h_{pair_{i2}}^{\bm{x}_1} \\
                    -L_{g}h_{pair_{i3}}^{\bm{x}_1} \\
                    \vdots \\
                    -L_{g}h_{pair_{iN_a}}^{\bm{x}_1}
                   \end{bmatrix} \bm{u}_i \leq 
                   \begin{bmatrix}
                    \zeta^{min}_2 + \alpha(h_{pair}(\bm{x}_1, \bm{x_2})) \\
                    \zeta^{min}_3 + \alpha(h_{pair}(\bm{x}_1, \bm{x_3})) \\
                    \vdots \\
                    \zeta^{min}_{N_a} + \alpha(h_{pair}(\bm{x}_1, \bm{x_{N_a}})) \\
                   \end{bmatrix} \nonumber
\end{align}

\section{MAIN RESULT: APPLICATION}\label{sec:implementation}
The practical application of the results in Section \ref{sec:theory} to the distributed multi-vehicle safety problem are described in this section.  

The first-order dynamics \eqref{eq:SI_dynamics} of the control state \eqref{eq:control_state} are
\begin{equation}
    \begin{bmatrix}
        \dot{x}_i \\
        \dot{y}_i \\
        \dot{\theta}_i
    \end{bmatrix} = \underbrace{\begin{bmatrix}
        cos(\theta_i)  & -\gamma sin(\theta_i) \\ sin(\theta_i) &  \gamma cos(\theta_i) \\ 0 & 1 
    \end{bmatrix}}_{g(\bm{x}_i)}
    \begin{bmatrix}
        u_{thr} \\
        u_{rud}
    \end{bmatrix}.
\end{equation}

The function
\begin{equation}
    h_{pair}(\bm{x}_i, \bm{x}_j) = (x_i - x_j)^2 + (y_i - y_j)^2 - r_{safe}^2,
\end{equation}
where $r_{safe}$ is the safety radius, defines the pair-wise safe set as 
\begin{equation}
    \mathcal{C} = \{ \bm{x}_i \in D \subset {\rm I\!R}^n :  (x_i - x_j)^2 + (y_i - y_j)^2 \geq r_{safe}^2\}.
\end{equation}
The constraints on the ego vehicle's control input arise from the Lie derivative of $h_{pair}(\bm{x}_i, \bm{x}_j)$ with respect to the ego vehicles state $\bm{x}_i$ along $g$.  Specifically, this term is expressed as
\begin{align}
    L_{g}h_{pair_{ij}}^{\bm{x}_i} = \begin{bmatrix} 2(x_i - x_j)\\ 2(y_i -y_j) \\ 0 \end{bmatrix}^T\begin{bmatrix}
        cos(\theta_i)  & -\gamma sin(\theta_i) \\ sin(\theta_i) &  \gamma cos(\theta_i) \\ 0 & 1 
    \end{bmatrix}.
\end{align}
The same constraint formulation applies to computing the worst-case input from the $j^{th}$ agent and the bound $\zeta_j^{min}$ in \eqref{eq:zeta_min_j}.   These values are computed locally by the ego agent at each iteration and used in the construction of the QP.

\subsection{Role of the control point}
The primary role of the control point is to ensure that with $\gamma > 0$, then $\frac{\partial h_{pair}(\bm{x}_i, \bm{x}_j)}{\partial u_{rud}} \neq 0$.  Moreover, the magnitude of $\gamma$ influences the relative weighting of $u_{rud}$ versus $u_{thr}$.   Therefore, as a design principle, 
\begin{equation}
0 < \gamma < r_{safe} 
\end{equation}

The magnitude of $\gamma$ also plays a role in satisfying Assumption \ref{assump:ith_control_set}. Specifically if  $ \gamma \cdot u_{rud_{max}} \geq \ u_{thr_{max}}$, and all other combinations of the thruster and actuator limits have the same relationship, then Assumption 1 is satisfied for all values of $\theta_i$ and $\theta_j$.  The intuition behind that assertion is that even when a neighbor vehicle is approaching the ego vehicle broadside, the ego vehicle can rotate, moving the control point away from the worst case movement of the neighbors control point. 
Finally, using $\gamma$ is practical because it can be difficult to identify the actual pivot point of a surface vehicle. 

\subsection{Addition of slack variables}
In practice the nominal QP is sometimes not feasible and thus $K_{cbf}$ is empty.  These situations can happen upon initial deployment if vehicles are too close to each other and deadlock, or if the model of the dynamics do not adequately capture the complex motion of vehicles in the water and vehicles drift closer due to unmodeled forces such as wind and currents.  It is important to address this issue, and a common method is including a slack variable to dilate $\mathcal{C}$ just enough that $K_{cbf}$ is not empty, and the ego vehicle can execute a best effort control action.  

Slack variables are added to only the constraints that cause the QP to not be feasible, ensuring that dilation of $\mathcal{C}$ only occurs when absolutely necessary.  The slack variables are given a high penalty in $\bm{Q}$, incentivizing the solver to find a solution with aggressive control action that will rapidly move the vehicle up the level-sets of $h_{pair}$ towards $\text{Int} (\mathcal{C})$.  


\section{EXPERIMENTAL SETUP}
\begin{figure}[ht]
    \centering
    \includegraphics[width=1.0\linewidth]{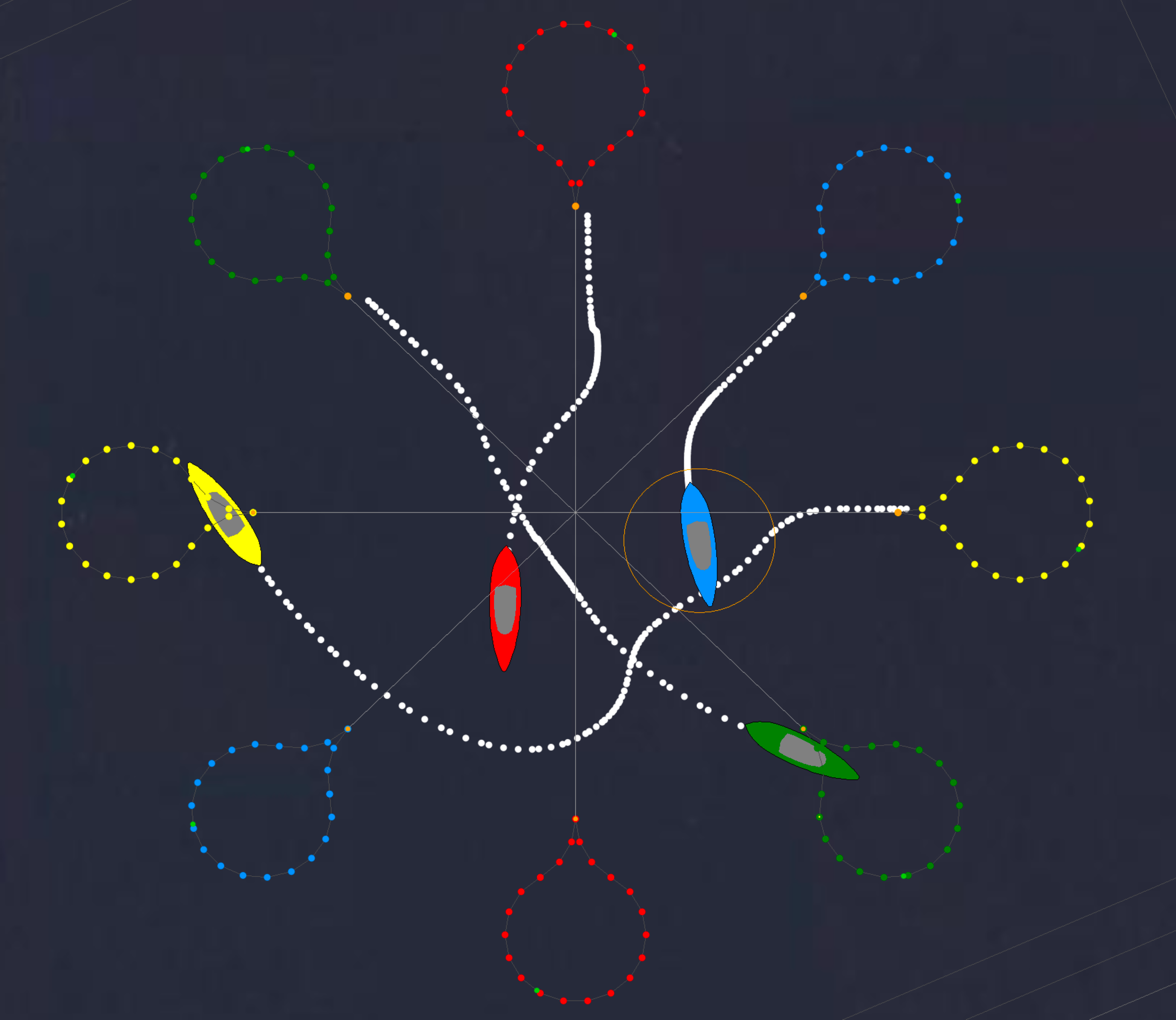}
    \caption{Overview of the joust mission used for structured evaluation of safety.  Each vehicle starts at one side of a 64 meter circle and at the same time they all begin to travel to their own goal point on the other side of the circle, maneuvering around the other vehicles in their way.  Once they have reached the other side, each vehicle executes a 360 degree turn to prepare for the next cycle. }
    \label{fig:joust_mission}
    \vspace{-3mm}
\end{figure}

The joust mission is common and structured way to evaluate the performance of collision avoidance in multi-vehicle situations.  An example of the experiment is shown in Figure \ref{fig:joust_mission}.  This same experiment has been used to test new approaches to collision avoidance \cite{molina2024methods, Martinez2025AVOCADO} including CBF-based approaches \cite{Wang2017CBFTRO}.  It is intended to specifically stress the decision-making and control execution components that are needed for collision avoidance.

Vehicle speed, or more precisely the relative speed between two vehicles, plays a significant role in how vehicles maneuver to avoid colliding with each other.  For this reason the nominal speed of each vehicle in the joust mission is (uniformly) randomly set between 1.0 and 2.0 meters per second, the typical operating speed for the two USVs used for evaluation, and this value was reset every 100 seconds.  Each vehicle can adjust speed to avoid collisions using the COLREGS behavior, the CBF filter, or both.

Another factor in maneuvering is the ability of vehicles to turn, particularity at low speeds and small $u_{thr}$ values.  In many vehicles the ability to rotate, or drive $\dot{\theta}_i$, is diminished when thruster input is small.  To capture this effect on the differential thrust USVs used in experimentation, which could theoretically rotate in place, a thruster mapping is used in both the simulated studies and the real-world experiments.   In addition, the same restriction is included in the simulated vehicle dynamics.   It is important to emphasize that the ego agent assumes every neighbor can essentially rotate in place while computing its worst-case assessment of the pairwise CBF, which is clearly a conservative assumption from a safety perspective.

\begin{table}[ht]
\caption{Key parameters used in simulation and experiments}
\begin{tabular}{ccccc}
CBF Parameter  & Value & & COLREGs Bhv. Parameter & Value \\ \cline{1-2} \cline{4-5} 
$r_{safe}$      & 15 m            & & pwt\_outer\_dist      &  30 m \\
$\gamma$        & 2 m             & & pwt\_inner\_dist      &  20 m \\
$\alpha(h)$     & $1.0 \cdot h$   & & min\_util\_cpa\_dist  &  10 m \\
                &                 & & max\_util\_cpa\_dist  &  20 m 
\end{tabular}
   \vspace{-5mm}
\end{table}

\subsection{Hardware}
\begin{figure}[ht]
    \centering
    \includegraphics[width=1.0\linewidth]{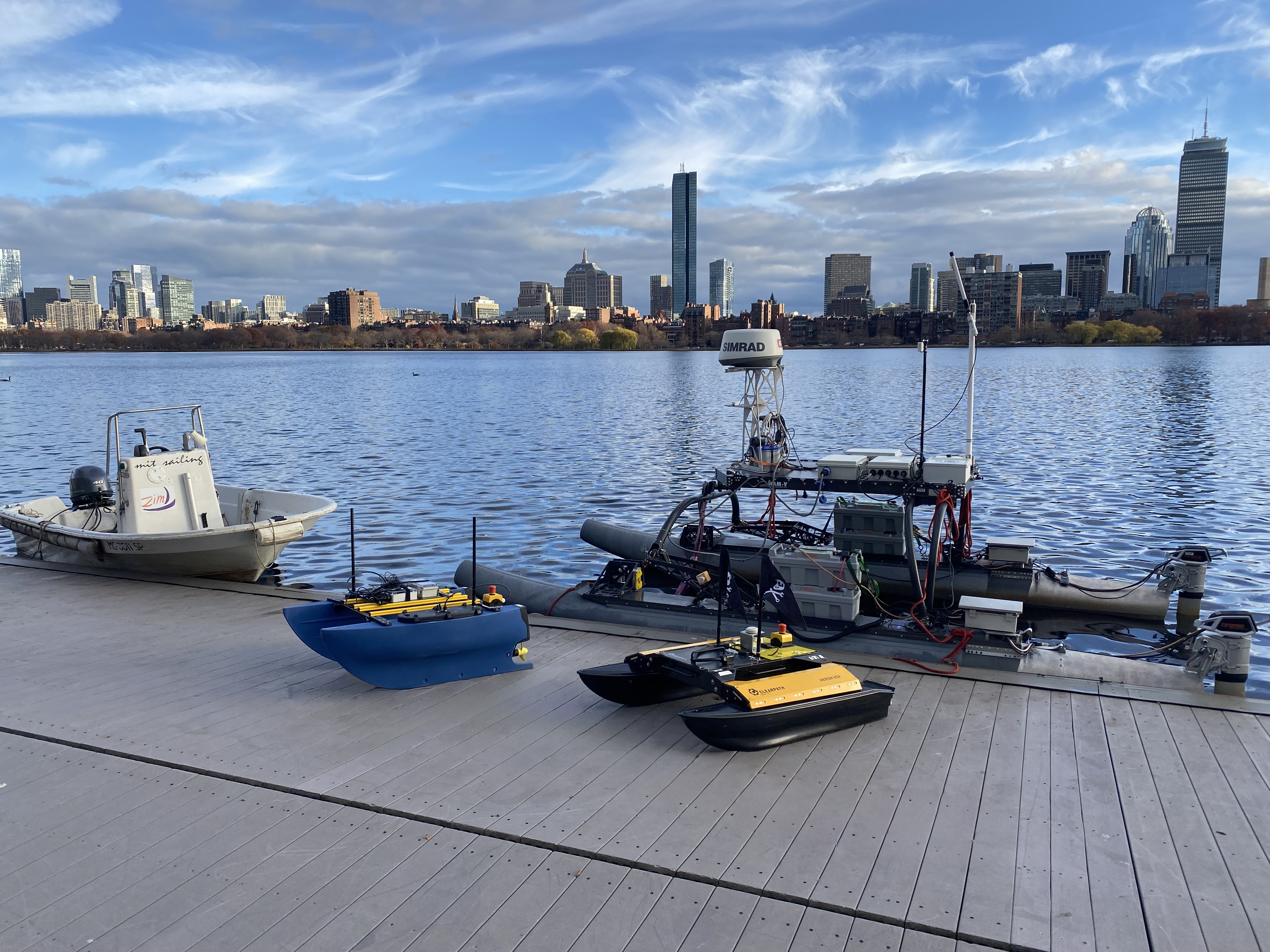}
    \caption{The four vehicles used in joust mission experiments. From left to right: a human operated small watercraft, a BlueBoat USV, a Heron USV, and a 16ft WAM-V USV. }
    \label{fig:four_vehicles}
    \vspace{-5mm}
\end{figure}

The hardware experiments involved three USVs: the Clearpath Heron USV, Blue Robotics BlueBoat, and the 16ft OPT WAM-V shown in Figure \ref{fig:four_vehicles}.  The fourth vehicle was a small human operated boat.  Although each vehicle used MOOS-IvP with the same behaviors, nominal PID controller, and CBF-based safety filter, each had a different set of dynamics, sensors, thrusters, PID parameters, and thruster allocation methods.

\section{SIMULATED RESULTS AND DISCUSSION}

\begin{table*}[ht]
\centering
\caption{Performance metrics in simulated encounters.  The data for each row are taken from 100,000 encounters.}
\label{tab:sim_res}
\begin{tabular}{|l|c|c|c|c|c|}
\hline
Metric:                 & \begin{tabular}[c]{@{}c@{}}Variation of encounter \\ frequency in angular \\ direction (metric of \\ simulation coverage)\end{tabular} & \begin{tabular}[c]{@{}c@{}}Near misses in \\ 100,000 encounters \\ (metric of safety)\end{tabular} & \begin{tabular}[c]{@{}c@{}}Collisions in \\ 100,000 encounters\\ (metric of safety)\end{tabular} & \begin{tabular}[c]{@{}c@{}}Average extra time \\ required per leg \\ (metric of efficiency)\end{tabular} & \begin{tabular}[c]{@{}c@{}}Average extra distance \\ traveled per leg \\ (metric of efficiency)\end{tabular} \\ \hline
Only COLREGS behaviors    & 0.17 &  929 & 34 & +22\%    &  \textbf{+3.6\%} \\ \hline
Only CBF                  & \textbf{0.10} &  174 & 0  & +24\%   &  +25\%  \\ \hline
\begin{tabular}[c]{@{}l@{}}COLREGS behaviors \\ + CBF\end{tabular}   & 0.19 & \textbf{13} &  \textbf{0} &  \textbf{+21\%} & +5.4\% \\ \hline
\end{tabular}
   \vspace{-3mm}
\end{table*}


The data on the simulated encounters in the joust mission are presented in Table \ref{tab:sim_res}.  Simulated external disturbances were present in these encounters, with uniformly random magnitude varying from 0.01 to 0.02 m/s in a uniformly random direction.  Due to the simulated wind and waves, as well as the random change in the nominal speed, each set of simulated trials included a variety of situations where a vehicle had to contend with one, two, or three simultaneous contacts approaching from arbitrary directions.  A more quantitative evaluation of the uniformity of the encounter randomization is covered in Section \ref{sec:sub_coverage}.

The safety metrics evaluated during each simulation were:
\begin{itemize}
    \item \textbf{Near misses:} The number of encounters where the minimum range between two vessels during the encounter was less than 10 meters.
    \item \textbf{Collisions:} The number of encounters where the minimum range between two vessels during the encounter was less than 3 meters.
\end{itemize}

The efficiency metrics were:
\begin{itemize}
    \item \textbf{Average extra time to complete the leg:} A percentage change over a baseline traversal across the circle with no encounters.
    \item \textbf{Average extra distance to complete the leg:} A percentage change over a baseline traversal across the circle with no encounters.
\end{itemize}

\subsection{Quantifying adequate randomization of encounter positions in simulation}\label{sec:sub_coverage}
\begin{figure}[ht]
    \centering
    \includegraphics[width=0.9\linewidth]{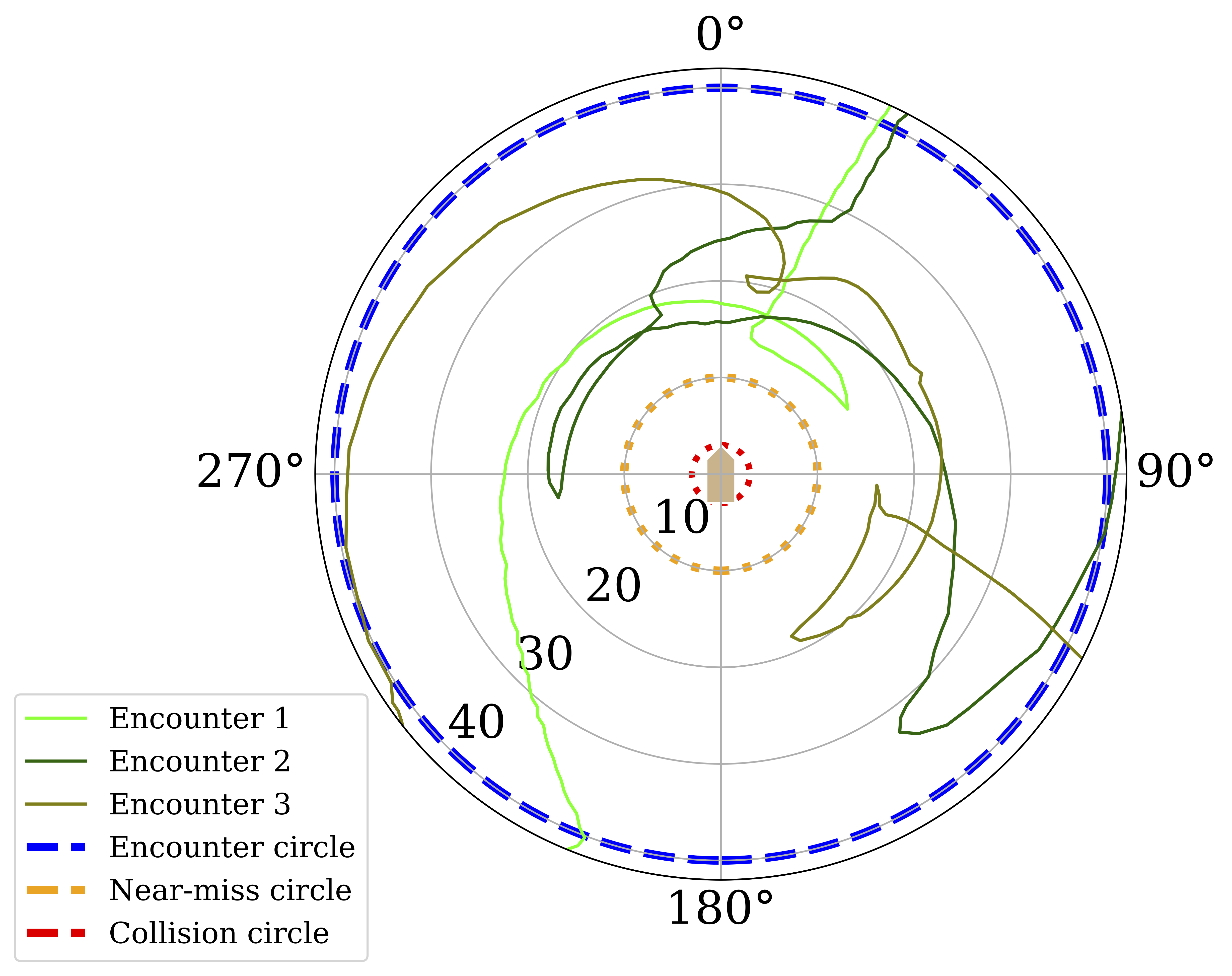}
    \caption{Trajectories of three simulated encounters (green) in the CBF-only trials.  Each encounter is plotted relative to the perspective of the ego vehicle.   Each radial axis label is in meters, and the radius of the collision circle is 3 meters.}
    \label{fig:sim_encouters}
\end{figure}

The trajectories of all randomized encounters are plotted relative to the position and orientation of the ego vehicle.  An example of three encounters are plotted in Figure \ref{fig:sim_encouters}.  

\begin{figure}[ht]
    \centering
    \includegraphics[width=0.9\linewidth]{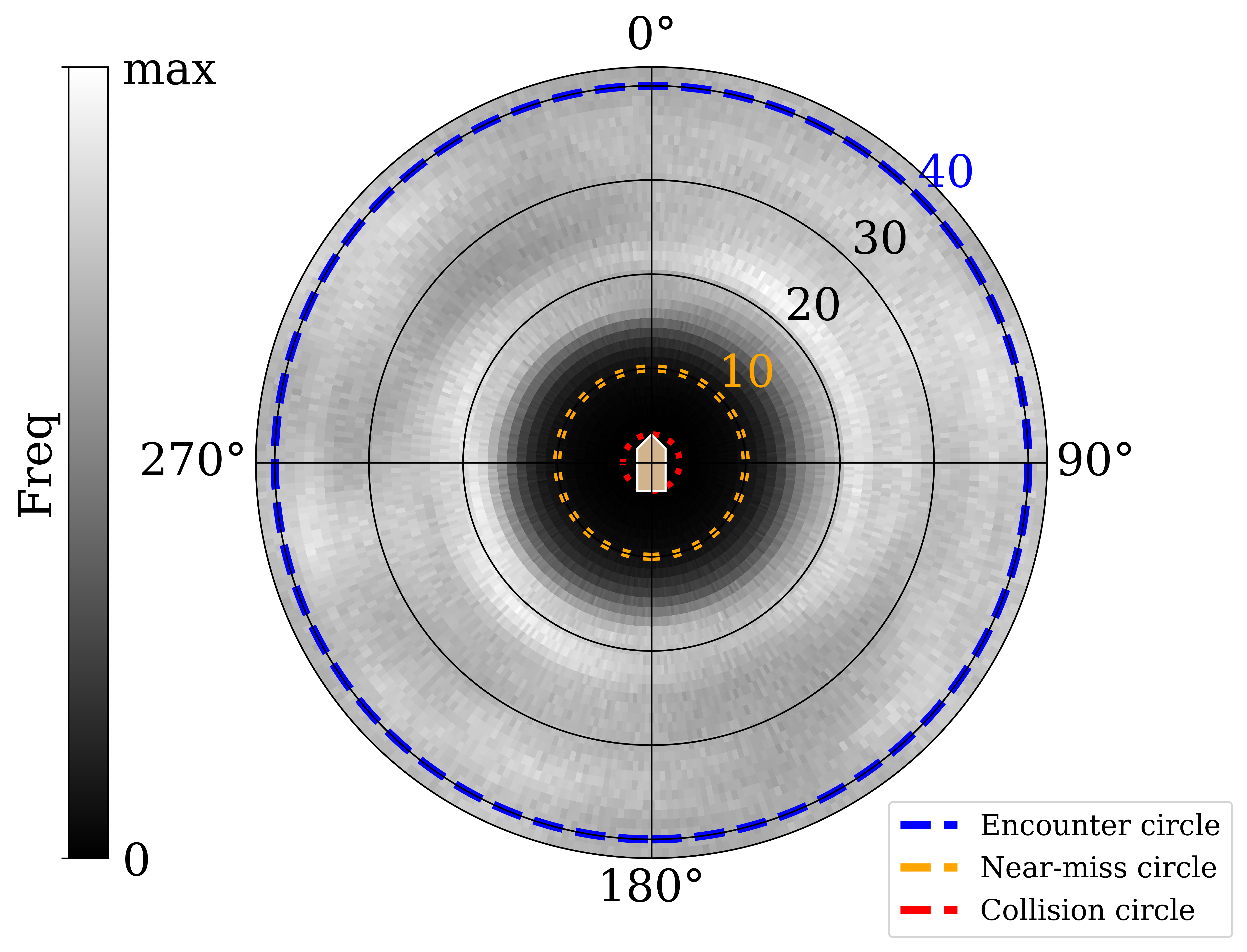}
    \caption{Frequency of the relative positions of contacts in 100,000 simulated encounters in the CBF only trials.  All encounters are plotted relative to the perspective of the ego vehicle. A metric that quantifies the uniformity of these encounters in the angular dimension is reported in Table \ref{tab:sim_res}.}
    \label{fig:encouter_freq}
       \vspace{-3mm}
\end{figure}

To compute the coverage metric in Table \ref{tab:sim_res} the relative position of all encounters are binned into a grid of ranges and bearings for a total of $2 \times 100,000$ entries, one for each view of the encounter.  The grid interval for range was $0.1m$ and $1$ degree for bearing.  The grid for the CBF-only experiments is shown in Figure \ref{fig:encouter_freq}.  

The radial profile of encounter frequency is defined as the normalized function $\beta_k(\bar{r}) \in [0,1]$, where the value of $\beta_k(\bar{r})$ is the number of times an encounter was present along the bearing $k$ at radius interval $\bar{r}$, normalized by the maximum encounters of all grid cells.  The average radial profile of the frequency plot is defined as $\mu_\beta(\bar{r})$, where $\mu_\beta(\bar{r})$ is the average number of encounters at that radius.  Finally, the variance metric is calculated as
\begin{equation}
    \mathbb{V}[\beta] = \frac{1}{360}\sum_{k=0}^{359}{\big|\big| \beta_k(\bar{r}) - \mu_{\beta}(\bar{r}})\big|\big|^2
\end{equation}
A value $\mathbb{V}[\beta] = 0$ implies the encounter frequency is asymmetrical.  The values for each set of simulated runs are provided in Table \ref{tab:sim_res}.

\subsection{Discussion}
Overall, the results indicate that the CBF-based safety filter eliminates collisions, and the COLREGS-based behavior improves efficiency in resolving encounters while also improving safety by reducing the near misses.   
In particular, all methods involving CBFs resulted in 0 collisions and drastically lower near-misses than the COLREGS behaviors alone. In addition, COLREGS + CBF methods were superior in both metrics of efficiency than the CBF method alone (21\% vs 24\% and 5.4\% vs 25\%). 
This suggests that the simultaneous use of both COLREGS-based behavior and CBFs results in the best blend of safety and performance. When augmented with CBFs, the COLREGS behaviors encouraged vehicles to reduce speed to let an encounter resolve while preserving safety, illustrating the important role of patience in both safety and efficiency. In addition, there was good coverage of all possible encounter ranges and bearings in the Monte-Carlo simulations, which is evidence of the generality of our proposed method.

\section{EXPERIMENTAL RESULTS}
\subsection{Joust mission experiments with four different vehicles}

\begin{figure}[ht]
    \centering
    \includegraphics[width=1.0\linewidth]{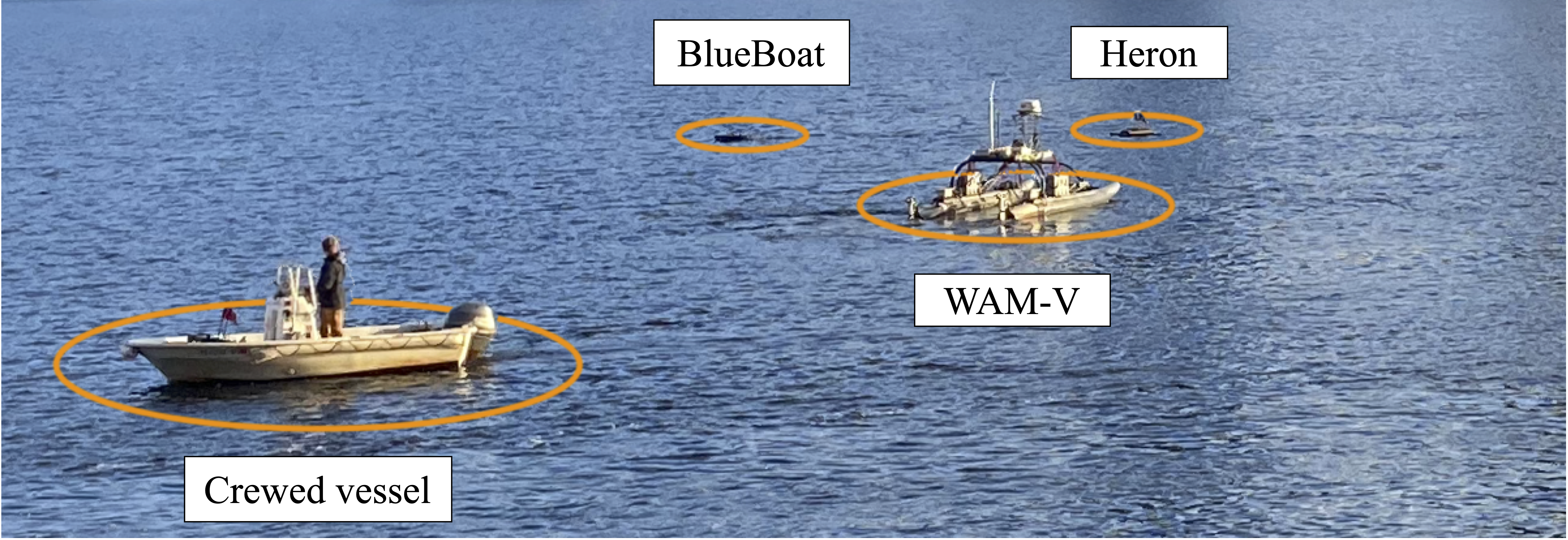}
    \caption{All four vehicles completing one round of the jousting mission.}
    \label{fig:CBF_Joust_experiment_photo}
       \vspace{-3mm}
\end{figure}

The four different vehicles shown in Figure \ref{fig:CBF_Joust_experiment_photo} successfully completed several rounds in the jousting mission over the course of an hour. 

\begin{figure}[ht]
    \centering
    \includegraphics[width=1.0\linewidth]{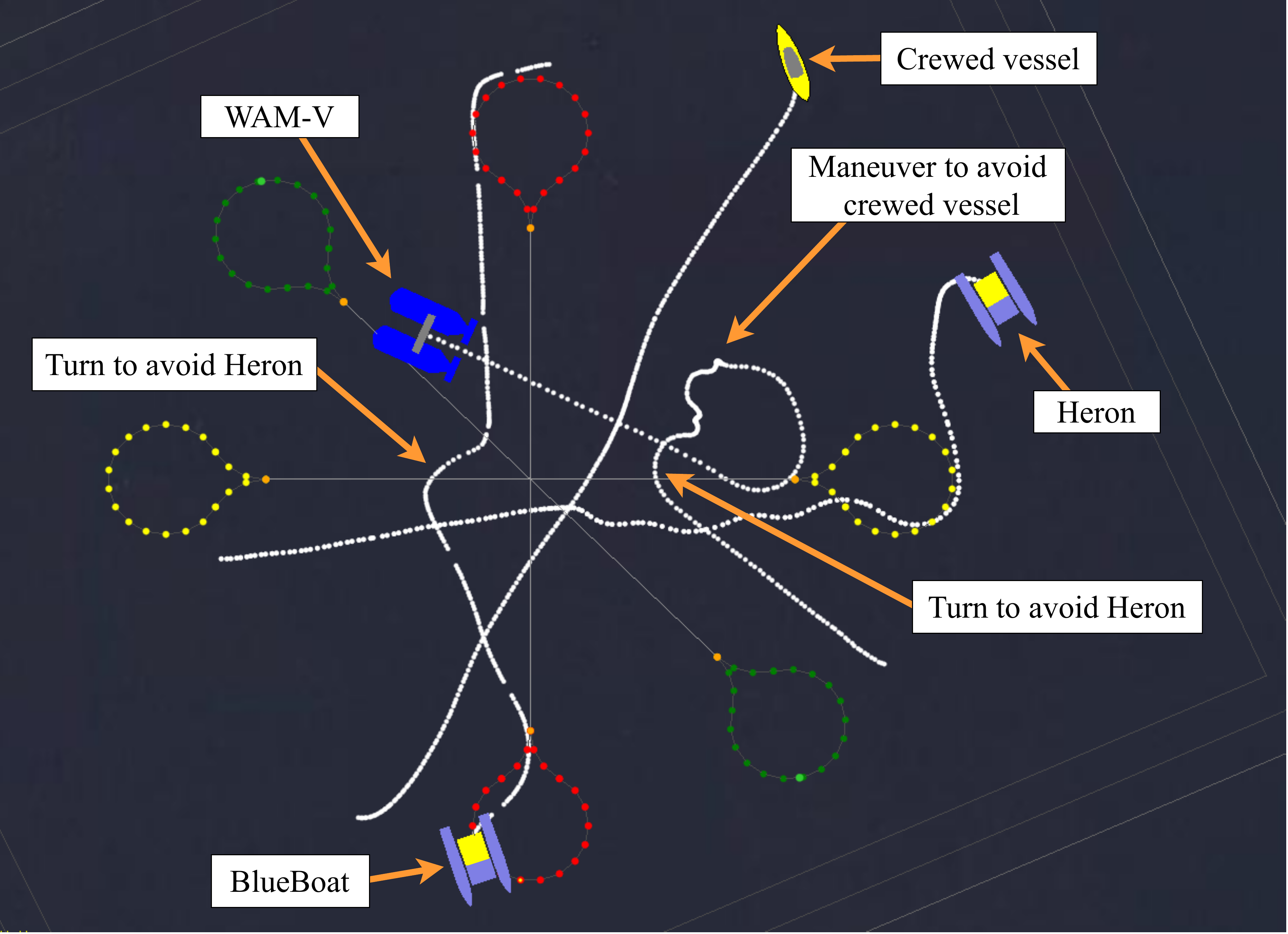}
    \caption{Vehicle trajectories for one round of the joust mission with four vehicles.  In this round only the CBF method was used for collision avoidance, and it was only active on the three autonomous vehicles.  All autonomous vehicles avoided each other and the crewed vessel.}
    \label{fig:CBF_Joust_experiment}
\end{figure}

The experiments were designed to highlight the robustness to adversarial contacts. In these experiments the human operated vessel purposefully did not always follow the COLREGS protocols, but instead attempted to drive as straight as possible to the other side.  In addition, the experiments were designed to evaluate the performance on autonomous vehicles that use vectored thrust to rotate, since these vehicles are more constrained in their movement than differential drive vehicles.  For this reason, the thruster allocation on the BlueBoat and the Heron were configured to mimic a vehicle that used vectored thrust, preventing them from utilizing the full capability of their differential thruster configuration.  For safety reasons, the largest vehicle, the 16ft WAM-V, was configured to use differential drive for better maneuverability.  A contrast between the maneuverability of these three vehicles can be seen in the trajectories of Figure \ref{fig:CBF_Joust_experiment}.   The WAM-V had much sharper turns than the other vehicles. Finally, a larger value of $20m$ for $r_{safe}$ was used for the larger WAM-V for additional safety.

\subsection{Discussion}
These hardware experiments demonstrate that our proposed method can successfully prevent collisions during complex multi-agent interactions involving heterogeneous vehicles. Our method enabled vehicles to navigate the many `in extremis' situations that occurred during the scenario.
Similar to the simulation studies, two sets of experiments were conducted, one with only the CBF active, and the other with the CBF and the COLREGS behaviors active.  
The experimental results reinforced the finding from simulation studies that while the CBF is necessary to prevent collisions, COLREGS behaviors improve efficiency in resolving encounters.  A good example of a situation where the CBF maintained safety, but the trajectory was not efficient is illustrated in Figure \ref{fig:CBF_Joust_experiment}, where the WAM-V makes several sharp turns before completely turning around.

\subsection{Validation in other multi-vehicle experiments}
The distributed CBF approach was used during other recent multi-vehicle experiments.  In addition to several trials of the joust mission with four Heron USVs, the distributed CBF was used to ensure safety when learned behaviors are fielded in large groups of USVs \cite{gonzalez2025ICRA}.  Figure \ref{fig:USV_fleet_exp} shows an aerial view of the experiments with learned behaviors featured in \cite{gonzalez2025ICRA}.  One qualitative conclusion from these experiments was that the human supervisors felt a reduction in stress during experiments. Overall, the positive results reinforced the merits of encapsulating complex or learned behaviors within a layer of control filtering for safety. 

\section{CONCLUSION} \label{sec:conc}
Autonomously avoiding collisions is essential to safely fielding large USV fleets, particularly when crewed vessels are participating alongside autonomous vehicles.  This study demonstrates that CBF-based safety filters can eliminate collisions in situations that are beyond the scope of COLREGS.   This method is a practical solution to this problem because it does not require vehicles to share information about intent or control.  Instead, each vehicle assumes the worst case behavior from all other contacts.  Finally, the results indicate that using a safety filter alone is not the best approach, since including a behavior based on the COLREGS protocol improves efficiency in resolving contacts.  

\bibliographystyle{IEEEtran}
\bibliography{IEEEabrv,refs}

\end{document}